# Flexible Policy Construction by Information Refinement

**Michael C. Horsch**
horsch@cs.ubc.ca

**David Poole**
poole@cs.ubc.ca

Department of Computer Science
University of British Columbia
2366 Main Mall,
Vancouver, B.C., Canada V6T 1Z4

## Abstract

We report on work towards flexible algorithms for solving decision problems represented as influence diagrams. An algorithm is given to construct a tree structure for each decision node in an influence diagram. Each tree represents a decision function and is constructed incrementally. The improvements to the tree converge to the optimal decision function (neglecting computational costs) and the asymptotic behaviour is only a constant factor worse than dynamic programming techniques, counting the number of Bayesian network queries. Empirical results show how expected utility increases with the size of the tree and the number of Bayesian net calculations.

## 1 INTRODUCTION

Influence diagrams provide expressive and intuitive representations for an important class of decision problems [Howard and Matheson, 1981; Shachter, 1986; Pearl, 1988]. Small problems can be solved by finding a policy which maximizes the decision maker's expected utility without considering the cost of computation, but finding such a policy requires an exponential number of maximizations, in terms of the number of decisions in the problem [Shachter, 1986; Shachter and Peot, 1992; Zhang, 1994; Cooper, 1990].

When the costs of computation are taken into account, the decision maker must be concerned not only about the outcomes of acting in the world, but also about effects of computing on finite hardware while other processes in the world continue [Horvitz, 1990; Russell and Wefald, 1992].

We report a technique to compute policies for decision problems expressed as influence diagrams. For each decision node in the influence diagram, the technique builds a decision function in the form of a tree whose non-leaf vertices are labelled with predecessors of the decision node, and whose leaf vertices are labelled with actions.

For each decision node in an influence diagram, our technique iteratively refines the information contained in the tree. The tree starts as a single leaf, representing an action to be carried out regardless of the available information. The incremental improvement replaces a leaf, *i.e.*, an action taken based on a certain subset of information, with a vertex, *i.e.*, a variable which is observable to the decision maker. New leaves are added, representing actions to be taken given the distinctions made in the path from root to leaf in the tree.

The decision function can be refined in this way until some stopping criterion is met, or until there are no more observable variables to add to the tree. A tree which cannot be further refined can represent the same "optimal" decision function that would be computed by traditional dynamic programming techniques.

For a decision node with $n$ informational predecessors, each having at most $b$ values, the sequence of improvements to a tree ends after $O(b^n)$ queries to a Bayesian network [Shachter and Peot, 1992], and is a constant factor worse than traditional dynamic programming techniques. However, our technique is not intended for finding optimal policies; the advantage of our technique is that a policy is available immediately, and it is improved incrementally with further computation.

This technique is a step towards flexible iterative refinement of policies for decision problems [Horvitz, 1990]: decision functions are available in an anytime manner, with an non-decreasing expected value as more computational resources are devoted to the problem and the next iteration in the sequence is performed by *refining* the previous policy. Each tree in the sequence represents a sub-optimal decision function, whose expected value to the decision maker is well defined.

Our empirical results suggest that small trees can be computed before an optimal decision function, and while they are sub-optimal, they have positive value for a decision maker reasoning with bounded resources.

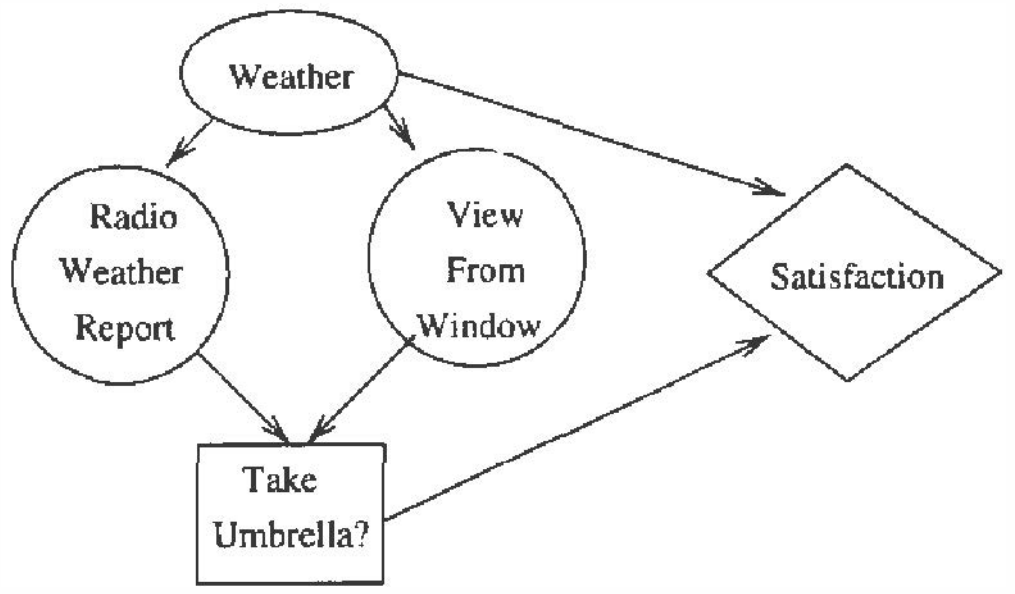

Figure 1: *A simple ID*

## 2 INFLUENCE DIAGRAMS

An influence diagram (ID) is a directed acyclic graph representing a sequential decision problem under uncertainty [Howard and Matheson, 1981]. An ID models the subjective beliefs, preferences, and available actions from the perspective of a single decision maker.

Nodes in an ID are of three types. Circle shaped *chance* nodes represent random variables which the decision maker cannot control, square shaped *decision* nodes represent decisions, *i.e.*, sets of mutually exclusive actions which the decision maker can take. The diamond shaped *value* node represents the decision maker's preferences.

Arcs represent dependencies. A chance node is conditionally independent of its non-descendants given its direct predecessors. A decision maker will observe a value for each of a decision node's direct predecessors before an action must be taken. The decision maker's preferences are expressed as a function of the value node's direct predecessors.

In an ID, there is a conditional probability table associated with every chance node (unconditional, if it has no predecessors), and a value function associated with the value node.

For example, Figure 1 shows an augmented version of the well known *Weather* ID [Shachter and Peot, 1992]. The ID represents the information relevant to a hypothetical decision maker, whose problem is to decide whether to take an umbrella to work. The goal is to maximize the decision maker's expected *Satisfaction*, which depends on the *Weather* and decision maker's decision to *Take Umbrella?* The decision maker can choose to *Bring Umbrella*, or *Leave Umbrella*, which are not explicit in the figure.

The decision maker has two sources of information: a *Radio Weather Report*, and the *View From Window*. These random variables are explicitly assumed to be independent given the weather, and both have three possible outcomes: *Sunny*, *Cloudy*, and *Rainy* (not explicit in the figure). The *Weather* is also a random variable, not directly observable at the time an action must be taken; it has two states: *Sun* and *Rain*, (not explicit in the figure).

For brevity, probability and utility information for this example has not been shown. However, conditional probability tables of the form P(*Weather*), P(*Radio Weather Report*| *Weather*), and P(*View From Window*|*Weather*) are necessary to complete the specification. The value function, *Satisfaction*(*Weather*, *Take Umbrella*) is also necessary.

A policy prescribes an action (or sequence of actions, if there are several decision nodes) for each possible combination of outcomes of the observable variables. In one of the possible policies for the above example, the decision maker always takes an umbrella, regardless of the information available. An optimal policy is the policy which maximizes the decision maker's expected *Satisfaction*, without regard to the cost of finding such a policy.

The goal of maximizing the decision maker's expected *Satisfaction* can be achieved by finding an optimal policy, if computational costs are assumed to be negligible. If computational costs are not negligible, the decision maker's expected utility might be maximized by a policy which is not optimal in the above sense.

In this paper, IDs are assumed to have chance and decision nodes with a finite number of discrete values. Furthermore, we limit the discussion to IDs with a single value node.

### 2.1 NOTATION

Chance nodes are labelled $x, y, z, \ldots$. Decision nodes are labelled $d$, with subscripts if necessary to indicate the order the decision nodes. The value node, and its value function, will be labelled $v$.

The set of a node's direct predecessors is specified by $\Pi$ subscripted by the node's label. The set of values (outcomes or actions) which can be taken by a node is specified by $\Omega$, similarly subscripted. The set $\Omega_{\Pi_d}$ is the set of all possible combinations of values for decision node $d$'s direct predecessors. An element in this set will be called an *information state*.

A decision function for $d$ is a mapping $\delta : \Omega_{\Pi_d} \rightarrow \Omega_d$. A policy for an ID is a set $\Delta = \{\delta_i, i = 1 \ldots n\}$ of decision functions, one for each of the decision nodes $d_i, i = 1 \ldots n$.

### 2.2 RELATED WORK

There are several techniques for solving IDs, which do not consider the cost of computation. The original technique converts an ID to a symmetric decision tree [Howard and Matheson, 1981]. An algorithm which operates on the graphical structure is given in [Shachter, 1986].

Recent advances in efficient computation in Bayesian networks [Pearl, 1988; Lauritzen and Spiegelhalter, 1988; Jensen *et al.*, 1990] provides a framework for efficient computation of expected value and optimal policies [Shachter and Peot, 1992; Jensen *et al.*, 1994]. Heuristic search has also been applied to finding policies for IDs [Qi and Poole, 1995] using these advances. We use Bayesian networks (BNs) as the underlying computational engine for our technique to compute posterior probabilities and expected values [Shachter and Peot, 1992].

A number of researchers have described iterative approaches to solving influence diagrams. Heckerman *et al.*, [1989] and Lehner and Sadigh [1993] use tree structures to represent policies, and use a greedy approach to incremental improvement of the tree structure. Both approaches use a single tree to represent the policy. Lehner and Sadigh define optimality of a decision tree with respect to the number of nodes in the tree, and give a general property which guarantees that an optimal decision tree of a certain size can be found by greedy search. Heckerman *et al.* weigh the value of the tree against the cost of computing it, but the tree itself is intended as an alternative to on-line decision making.

Our work extends the previous work by building a decision function, in the form of a tree, for each decision node, taking advantage of efficient probabilistic inference techniques. This combination creates a basis for on-line, resource bounded computation.

# 3 SINGLE STAGE COMPUTATIONS

We use trees to represent decision functions. In this section we show how these trees can be built. We call the trees "decision trees" because of the relationship to the machine learning literature (*e.g.*, [Quinlan, 1986]). The reader should be aware that our decision trees differ from the operations research decision trees in that our decision trees are used to represent solutions (or partial solutions) to decision problems, whereas operations research decision trees represent a decision problem. However, the two have much similarity in structure, which seems sufficient reason to use a common name; the difference in their respective uses seems sufficient to prevent confusion.

In this section, we consider the case where the decision problem has a single decision node, and extend the idea to IDs with multiple decision nodes in Section 4.

## 3.1 DECISION TREES

Let $d$ be a decision node in an ID. A decision tree $t$ for $d$ is either a leaf labelled by an action $d_j \in \Omega_d$ or a non-leaf node labelled with some observable variable $x \in \Pi_d$. Each non-leaf has a child decision tree for every value $x_k \in \Omega_x$. An information predecessor $x \in \Pi_d$ appears at most once in any path from the root to a leaf. Each vertex $v$ has a *context*, $\gamma_v$, defined to be the conjunction of variable assignments on the path from the root of the tree to $v$. The action at the leaf represents the action to be taken in the context of the leaf. Given an information state $w \in \Omega_{\Pi_d}$, there is a corresponding path through a decision tree for $d$, starting at the root leading to a leaf, which is labelled with the prescribed action for $w$. Note that the context of an action need not contain every variable in $\Pi_d$.

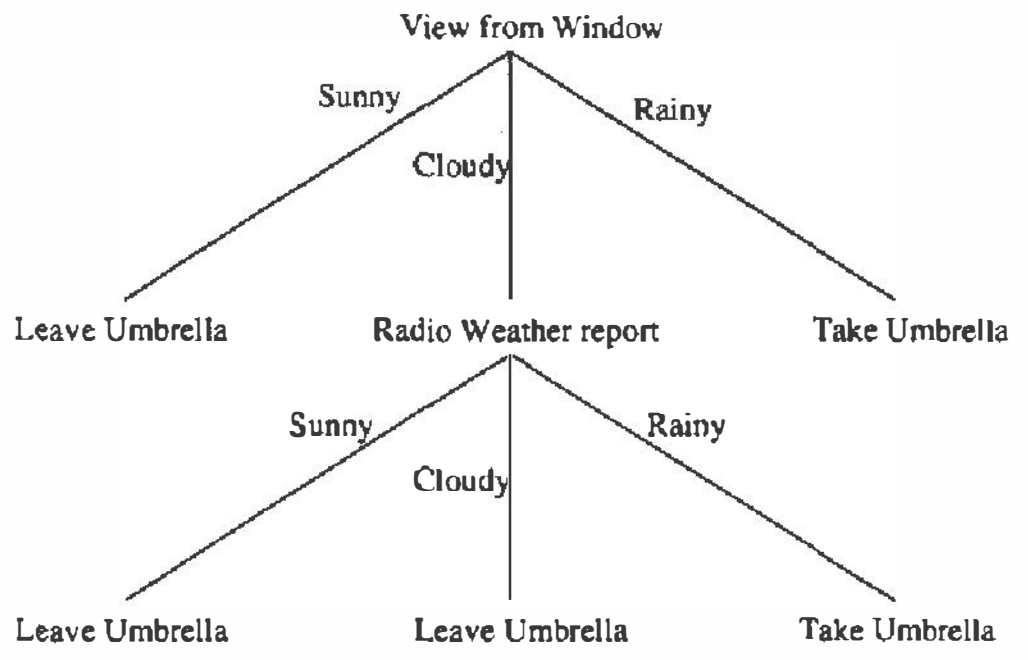

Figure 2: *A decision tree representation of a policy for the ID in Figure 1.*

A decision tree represents a decision function. We will refer to the action prescribed by a decision function by $\delta(w)$ for information state $w$, or by $d_l$ if $l$ is a leaf on a given decision tree.

A decision tree is shown in Figure 2. The tree can be interpreted as a policy for the ID in Figure 1 as follows. The decision maker first considers the view from the window. If the view is cloudy, then the decision maker will determine what to do by consulting the radio weather report. However, if the view from the window is sunny or rainy, then the radio report need not be consulted at all, even though the report is available as information. Note that in this example, the decision tree represents a policy. In general, we will construct a decision tree for each decision node in an influence diagram (Section 4).

We define expected value of a decision tree, so that we can compare decision trees, as follows:

$$E_t = \sum_{l \in t} u(d_l|\gamma_l)\mathrm{P}(\gamma_l)$$

where $u(d|\gamma)$ is the expected value of an action $d$ in a context $\gamma$, and the summation is over all leaves in $t$.

The optimal decision tree is defined as the one whose expected value is greater or equal to the expected value of any other decision tree: $t$ is optimal if for all $t'$, $E_t \geq E_{t'}$. This definition of optimal does not take into account the cost of computation.

## 3.2 THE SINGLE STAGE ALGORITHM

We build decision trees using a technique which resembles decision tree learning methods (*e.g.*, [Quinlan, 1986]).

For a given leaf $l$ in a decision tree, its context $\gamma_l$ is extensible if it does not contain all the observable variables. We refer to the variables which are not in the context as *possible extensions*, writing $\xi_l$. We will use $\gamma$ without subscript or argument when we need to refer to an arbitrary context. The symbol $\gamma_\phi$ represents the empty context, equivalent to the context at the root of a decision tree.

A decision tree $t$ can be extended if there is a leaf with an extensible context; otherwise, the tree is called *complete.* A tree is extended by removing leaf $l$ with context $\gamma_l$, and replacing it with new vertex $x \in \xi_l$. The issues of choosing a leaf to extend, and choosing a new vertex $x$ are discussed in Sections 3.4 and 3.3, respectively.

The new vertex $x$ is given a new leaf for every value $x_j \in \Omega_x$, and each leaf will be labelled with an action $d_j \in \Omega_d$ which maximizes the expected utility in the new context $\gamma' = x_j\gamma_l$.

The action at a leaf $l$ is chosen on the basis of being the best action for the context $\gamma_l$:

$$d_l = \arg \max_{d_i \in \Omega_d} u(d_i|\gamma_l)$$

This action can be determined with one query to a Bayesian network [Shachter and Peot, 1992] (see also Section 3.5). If a vertex has $b$ values, $b$ queries to the BN are required to compute its leaves. Another $b$ queries are needed to compute expected value for each leaf.

For a decision tree $t$, and leaf node $l$, we define the expected value of improvement, $EVI_t(l, x)$, to be the increase in expected value when $t$ is extended at $l$ with some $x \in \xi_l$, resulting in a new tree $t'$:

$$EVI_t(l, x) = E_{t'} - E_t$$

The basic algorithm can be given as follows:

```
procedure DT1
  Input:
    influence diagram ID
    with decision node d
  Output:
    a decision tree for d

  Start with the tree as a single leaf
  Do {
    Choose a leaf in the tree to extend
    Replace the leaf with an extension
  } Until   (stopping criteria are met
             or tree is complete)
  Return the tree
```

The sequence of trees created by DT1 is such that the expected utility of the next tree is never less than that of the previous tree. However, because the procedure is myopic (only a single vertex is added to a context at any time), there is no guarantee that the expected utility will always increase with every extension of the tree. In fact, an ID could be constructed in which the expected utility of a tree is arbitrarily far from the expected utility of the optimal tree as long as every leaf node is still extensible.

Note also that every decision tree in the sequence can be used by the decision maker at the time a decision must be made. Thus DT1 is an any-time algorithm. In this paper, we do not discuss stopping criteria for the algorithm.

The next two sections discuss in detail the issues of choosing a leaf to extend (Section 3.4) and choosing an extension for a given leaf (Section 3.3). These two topics are orthogonal.

### 3.3 CHOOSING AN EXTENSION FOR A GIVEN LEAF

A given leaf with an extensible context can be extended by choosing an informational variable which is not already in the context. In this section we discuss a greedy strategy.

The *best extension of t at l* is defined as follows:

$$\begin{aligned} x_l^* &= \arg \max_{x \in \xi_l} EVI_t(l, x) \\ &= \arg \max_{x \in \xi_l} \sum_{x_j \in \Omega_x} u(d_j|x_j\gamma_l)\mathrm{P}(x_j|\gamma_l) \end{aligned}$$

This $x_l^*$ represents the greedy improvement to the decision tree. Using this strategy for extending a leaf in DT1 results in a procedure we refer to as *Greedy*-DT1. In the above equation, $d_j$ refers to the action which maximizes the decision maker's expected utility with respect to the context $x_j\gamma_l$ (as described in the previous section).

Note that $EVI_t$ only depends on the leaf $l$, its context $\gamma_l$, and the possible extensions $\xi_l$. Therefore we can look at improvements to each leaf independently of possible improvements to other leaves, a property which we exploit computationally.

By choosing the best of all possible extensions for a given leaf, the *Greedy*-DT1 approach evaluates many extensions which it will never use, with the concomitant queries to the BN. However, the following results give a bound on the number of BN queries.

**Theorem 1** *Let $n$ be the number of information predecessors for a decision node $d$ in an ID. Furthermore, assume that the number of states for any node in the ID is bounded by a constant $b$. The total number of BN queries made by* Greedy–DT1 *in constructing a complete decision tree for $d$ is less than $\frac{2b}{(b-1)^2}b^{n+1}$.*

This follows from the observation that the number of extensions considered by *Greedy*–DT1 at a leaf depends on the size of its context: if decision node $d$ has $n$ predecessors, and a leaf has $k$ of these predecessors in its context, then the number of extensions which must be examined to choose the maximum is $n-k$. If the number of states a predecessor can take is bounded by constant $b$, then during the course of constructing the optimal decision tree, DT1 extends $b^k$ leaf nodes whose context is of size $k$. Therefore the total number of extensions considered by *Greedy*–DT1 is

$$\sum_{k=0}^{n-1}(n-k)b^k = b^{n-1}\sum_{k=0}^{n-1}(n-k)b^{k-n+1}$$

$$= b^{n-1} \sum_{i=1}^{n-1} i b^{1-i}$$
$$\leq b^{n-1} \sum_{i=1}^{\infty} i b^{1-i}$$
$$= b^{n-1} \left(1 - \frac{1}{b}\right)^{-2}$$

Finally, recall that each extension requires $2b$ queries to the BN (Section 3.2).

This result implies that *Greedy*–DT1 requires a constant factor of $\frac{2b^2}{(b-1)^2}$ more BN queries than [Shachter and Peot, 1992] to compute the complete tree. We emphasize that its advantage comes from the fact that it is an anytime algorithm.

However, choosing the best extension for a given leaf incurs a significant computational cost for small contexts. In order to heighten the any-time properties of this approach, we have explored the use of random extensions to a given leaf. This strategy, as we will see in Section 5 does not put as much effort into choosing how to extend a given leaf. Section 5 will show how a random approach performs, in comparison to *Greedy*–DT1.

## 3.4 CHOOSING A LEAF TO EXTEND

The DT1 algorithm does not specify how to choose a leaf to extend. Given the myopic nature of DT1, an obvious strategy would be to choose the leaf whose myopic extension would yield the highest increase in $EVI$. However, the increase in $EVI$ can only be determined exactly after an extension has been made. Thus some heuristic is necessary to determine which leaf node to extend.

We have explored several strategies, including choosing random leaves, and a *post hoc* heuristic, described below. These strategies do not make any assumption with respect to the way a particular leaf is extended. That is, the discussion in this section is independent of the issue presented in Section 3.3, where we discussed the best extension for a given leaf.

The effect of the post hoc heuristic is to spend computational effort in contexts where previous effort has provided the best previous results: A leaf is extended based on the increase in expected value which was gained when its parent vertex was added to the tree.

Descriptively, suppose a leaf $l$ is to be extended by adding a vertex $x$. New leaves are generated for $x$, by choosing actions which maximize expected utility for the new context which now includes $x$ (as described in Section 3.2). The extension brings an increase (possibly zero) in expected value to the tree. Likewise, the parent vertex of every leaf in the tree was added to the tree, increasing the tree's expected value. Of all such vertices, the post hoc heuristic chooses the one which brought the highest increase, and extends each of its leaves; ties are broken at random.

This heuristic has the advantage of using value information which is already known, as opposed to making an effort to estimate value information. As well, we note that, while the value of past effort doesn't always suggest a good place to make future effort, the incremental nature of DT1 will limit the amount of effort spent in any step.

We note that the post hoc heuristic requires a small amount of overhead ordering the most recent extensions to the tree. An alternative, involving less overhead, is to choose a leaf at random.

## 3.5 IMPLEMENTATION DETAILS

We have implemented the algorithm using two approaches: the *Greedy*-DT1 with the post-hoc heuristic, and a random approach, where leafs are chosen and extended at random. We use a BN to compute expected value, and make some effort to keep the implementation reasonably efficient.

### 3.5.1 The computational engine

Computation of expected value is based on the conversion of IDs to BNs [Shachter and Peot, 1992]. The utility node is converted to a binary chance node whose conditional probability distribution is the normalized utility function. Decision nodes in the ID are converted to root chance nodes with uniform probability distributions.

The BN derived in this way from an ID is subsequently compiled into a join tree, which can compute posterior probabilities efficiently, using DistributeEvidence and CollectEvidence operations [Jensen *et al.*, 1990].

Converting decision nodes to root chance nodes does not affect the probability distribution, and makes the size of the largest cliques in the join tree a function of the probabilistic information in the ID. The informational arcs remain part of the ID, but not the underlying BN.

The best action $d^* \in \Omega_d$ to be performed in a state $w \in \Omega_{\Pi_d}$ can be found by choosing the action which maximizes the query $P(d|w, v)$; the expected value of the best action is computed by querying the utility node $P(v|w, d^*)$.

We generalize this result in terms of choosing an action which maximizes expected utility in a given context $\gamma$ (recall that a context may not include a value for every observable variable in $\Pi_d$). The corresponding queries are: $P(d|v, \gamma)$, and $P(v|d^*, \gamma)$.

### 3.5.2 Efficiency issues

We use a priority queue to order the sequence of extensions to a given tree for the *post hoc* heuristic.

We avoid recomputation by storing the expected value of a leaf, and the expected value of each previous extension in the queue. As well, we store the posterior probability each of non-leaf node given its context. Thus, we need only to compute expected value for a new context; all the remaining information is obtained by look up.

In order to compute $EVI_t(l, x)$, we need to compute the posterior probability $P(x_j|\gamma_t(l))$. We order the exploration of extensions so that we can enter $\gamma_t(l)$ as findings once, and query the BN for the posterior probabilities of $x \in \xi_l$ on the basis of one DistributeEvidence computation. We still require this computation once for every leaf, but this is insignificant compared to the $2k$ CollectEvidence computations required to compute the expected value for all $k$ of the leaf's possible extensions.

In our implementation, any leaf node whose context contains logical impossibilities (*i.e.*, the context has probability zero) is not extended nor is it considered further. Empirically, this could have a great effect on the cost of computation and the size of the resulting tree, as shown in the first example in Section 5.

# 4 MULTISTAGE IDS

In this section, we present our preliminary work on extending the approach to IDs with multiple decision nodes; we will discuss future directions in Section 6.

We apply the DT1 algorithm in the traditional dynamic programming sequence: starting from the last decision in network, we compute a decision function, and then proceed backwards to the immediately preceding decision. We allow the contexts for a given decision tree to contain predecessor decision nodes (or not) according to the $EVI_t$.

We note that during the construction of a decision tree for decision node $d_k$, there are decision functions available for decision nodes $d_{k+1}, \ldots, d_n$, but not for $d_1, \ldots d_{k-1}$. The complication is that contexts for $d_k$ may not contain actions for all $d_1, \ldots d_{k-1}$, and furthermore, the decision functions for $d_{k+1}, \ldots, d_n$ may not contain actions for $d_1, \ldots d_k$.

To deal with this problem of incomplete contexts, we assume a uniform probability distribution over any action which is not mentioned in a given context. For an incomplete context $\gamma$, the process maximizes expected utility *as if* the decision maker would act randomly for any decision node not mentioned in $\gamma$; the policy, however, would not be stochastic in any sense.

The implementation applies DT1 to each decision node in sequence from last to first. When DT1 stops working on a particular decision node, the following steps are taken before the process can be applied to the immediately preceding decision node. First, some of the informational predecessors for the decision node are re-connected to the chance node which represents it using probabilistic arcs; only those predecessors which are used in the decision function are reconnected. A contingency table is then created which is consistent with the decision function, after the manner of [Shachter and Peot, 1992].

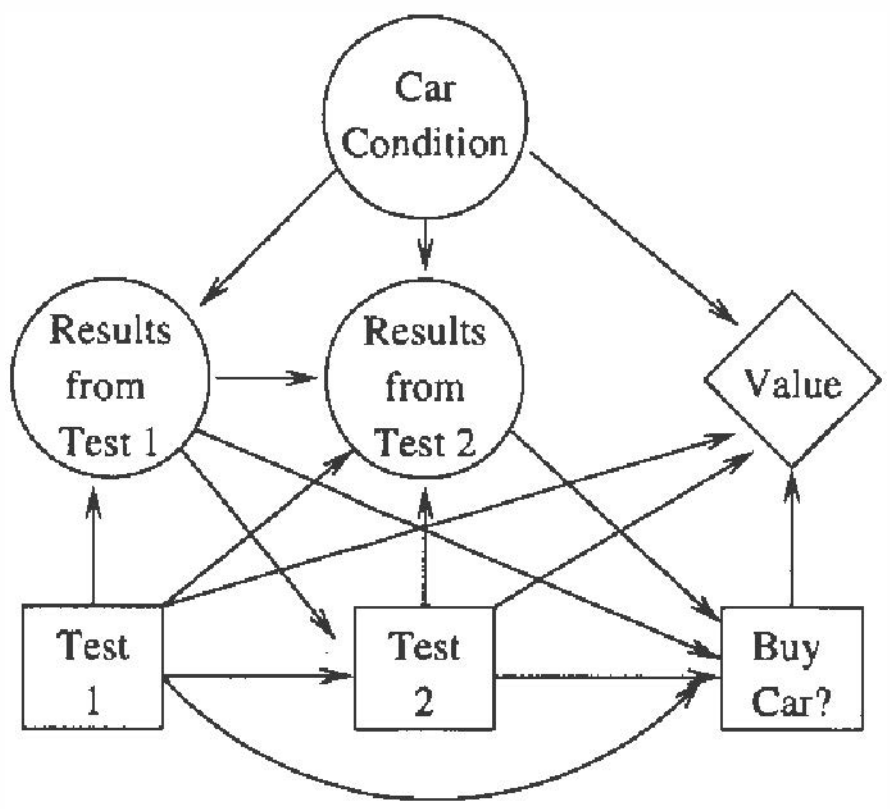

Figure 3: *The Car Buyer Problem, expressed as an ID.*

Note that the anytime properties of our approach are lost when DT1 is applied to decision nodes in sequence. This is due to the fact that the sweep back technique builds a policy in a fixed sequence, and once a decision tree is in place for $d_k$, any future extension to $d_k$ would invalidate the decision trees for decision nodes $d_1, \ldots, d_{k-1}$. Therefore, a multistage policy cannot be incrementally improved using the simple sweep back approach.

# 5 PERFORMANCE

In this section, we demonstrate the empirical behaviour of the *Greedy*-DT1 using the *post hoc* heuristic. We apply the algorithm to a small well known problem, and a set of larger random problems.

## 5.1 TERMS OF COMPARISON

We describe the behaviour of an implementation of DT1, running the procedure until the optimal decision tree is achieved. The data points we collect represent decision trees in terms of the tree's expected value, the number of BN computations required to achieve the tree, and the number of interior vertices in the tree. Note that each data point represents a decision function could be used as an anytime solution.

In our examples, value is normalized to [0, 1]. We measure the computation cost in terms of the number of BN computations required. Recall that each extensible leaf requires $2(n-k)+1$ BN computations, where $n$ is the number of predecessors of the decision node, and $k$ is the number of observations in the context of the leaf.

The size of our decision trees is measured in terms of the number of non-leaf vertices in the tree. Assuming binary valued predecessors of a decision node, the number of actions in a tree of size $s$ is $s+1$.

We compare our implementation to an hypothetical dynamic programming algorithm, in which a BN computation is required for every possible observable state [Shachter and

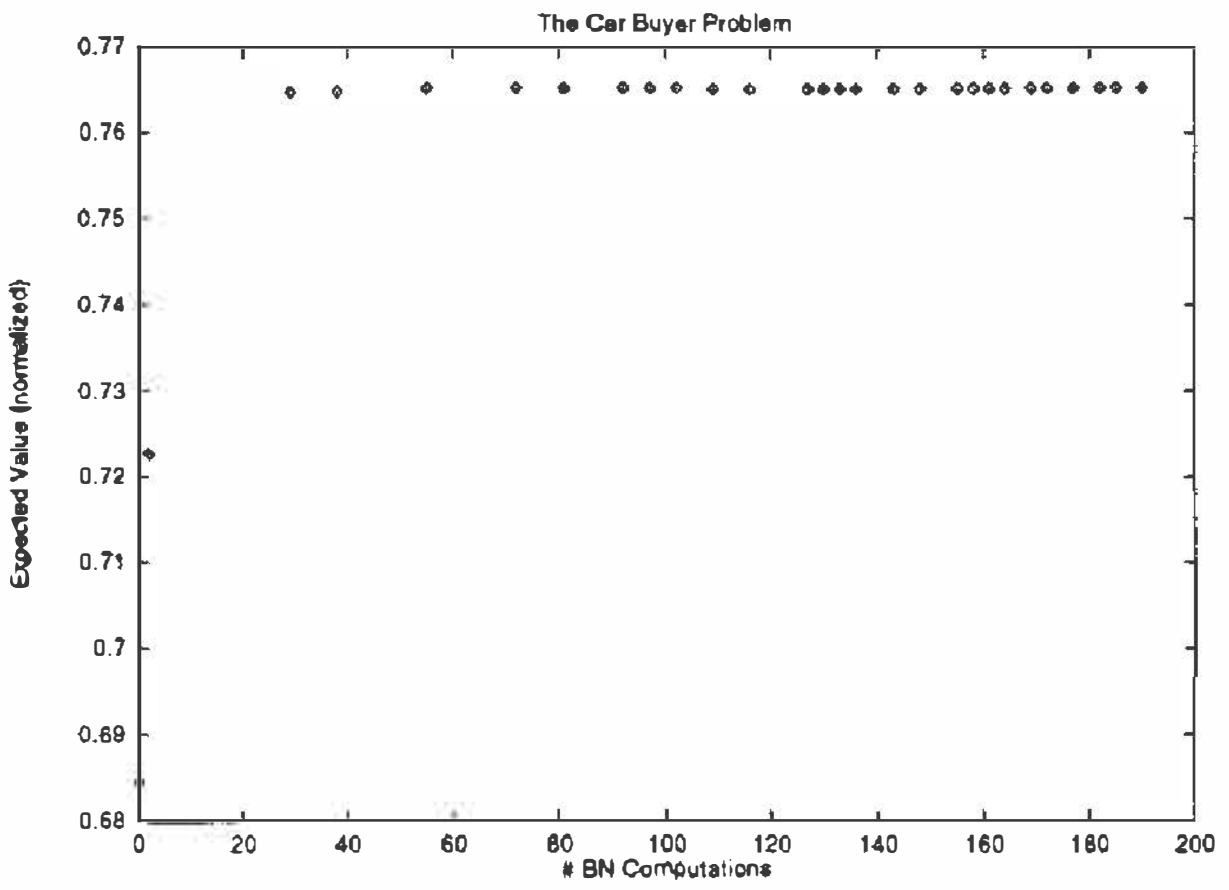

Figure 4: *The results of using* DT1 *on last decision node of the Car Buyer Problem. The expected value versus the number of BN computations. Each point represents a decision tree, and the rightmost point represents the optimal decision function.*

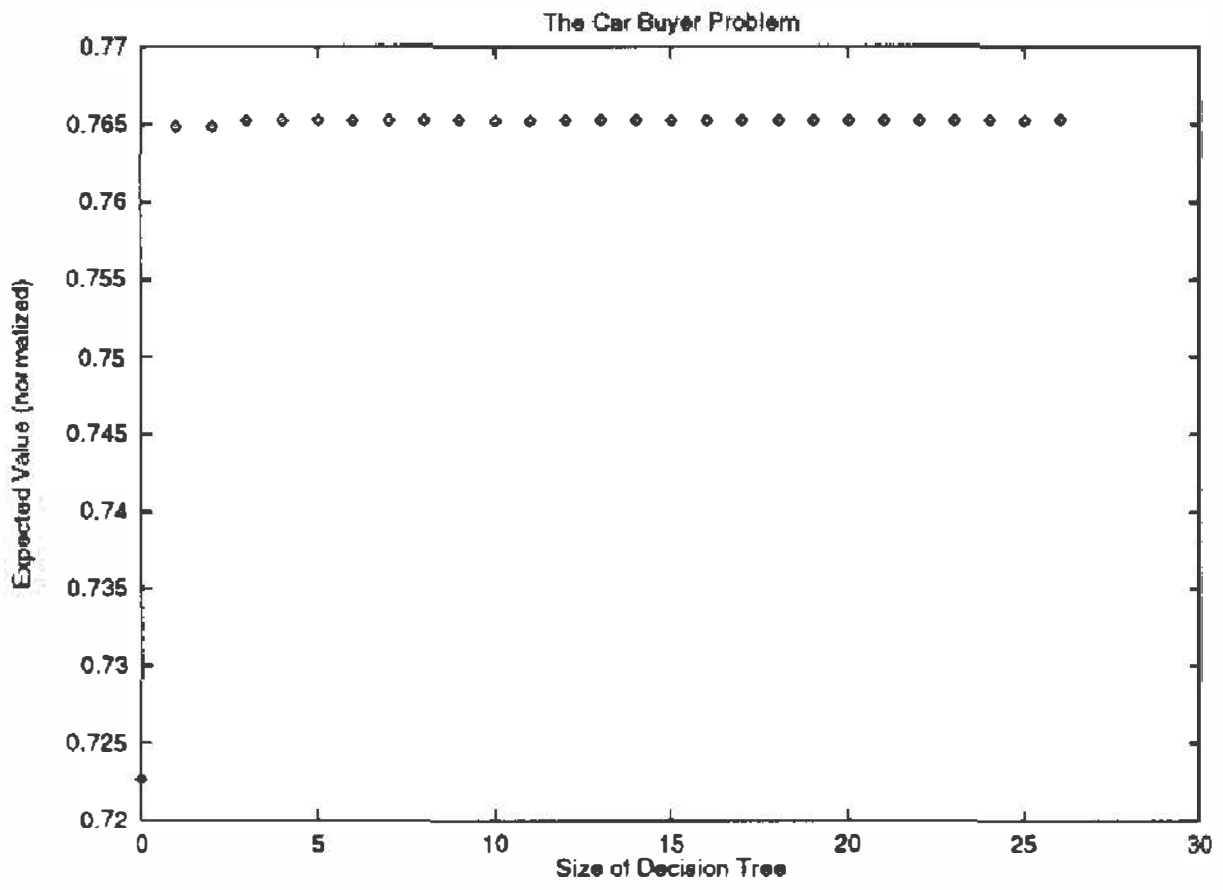

Figure 5: *The results of using* DT1 *on last decision node of the Car Buyer Problem. The expected value versus the size of the tree.*

Peot, 1992]. If a decision node in an ID has $n$ binary predecessors, then the dynamic programming algorithm requires exactly $2^n$ BN computations to find the optimal decision function containing $2^n$ information states. If required, the expected value of the decision function could be computed with a single BN computation after the decision function has been established. Finally, since the dynamic programming algorithm only produces a single solution, it falls under the category of *inflexible*, and for emphasis and brevity, we refer to it as such.

## 5.2 A CLASSICAL EXAMPLE

To illustrate the behaviour of DT1, we will show its behaviour on the well-known Car Buyer problem, Figure 3 [Smith *et al.*, 1993]. There are three decision nodes in this problem; in this section, we use DT1 to find a decision function for the last decision node.

The ID represents the knowledge relevant to a decision maker deciding whether or not to buy a particular car. The decision maker has the option of performing a number of tests to various components of the car, and the results of these tests will provide information to the decision to buy the car. The actual condition of the car is not observable directly at the time the decision maker must act, but will influence the final value of the possible transaction. A policy for this problem would indicate which tests to do under which circumstances, as well as a prescription to buy the car (or not) given the results of the tests. Due to space constraints, none of the numerical data required to complete the specification of this problem is shown; this information can be found in [Qi and Poole, 1995; Smith *et al.*, 1993]. This problem is well known for its asymmetry; some combinations of tests and results are logical impossibilities.

Figures 4, 5 and 6 show the performance of DT1 on the last decision stage in the problem.

Figure 5 compares the expected value of the decision tree and the size. Again, each point represents a decision tree improved by a single extension. Because of the asymmetries in the problem, the optimal decision function can be represented with a decision tree with 7 internal vertices and 10 leaves. For this problem, the largest tree computed by DT1 has 13 internal vertices and 13 leaves. The inflexible algorithm, unless designed to handle asymmetries (*e.g.*, [Qi and Poole, 1995]), requires a table of 96 entries. The large difference between our trees and the inflexible solution is due to the asymmetries in the problem; our implementation does not extend contexts whose probability is zero.

Figure 4 illustrates the increase in expected value of each decision function, as a function of the number of BN computations. The very first decision function for this problem is available after just 2 BN calculations, and has an expected value which is less than 10% from optimal. Each subsequent point represents an improvement to the decision function by extending the tree by one node. The rightmost point represents the complete decision tree.

For comparison, the inflexible algorithm requires 96 BN computations, to compute a table of 96 actions.

Figure 6 plots expected value versus the computational effort required; essentially, it shows where the work gets done. More work goes into finding the first few trees, as each leaf's context is small, and has many possible extensions. Half of the work is done to build a decision tree with 7 internal vertices. As the decision tree gets larger, the number of possible extensions for each context gets smaller. Towards the end of the process, fewer BN calculations are performed because there are no extensions to the contexts pulled from the queue.

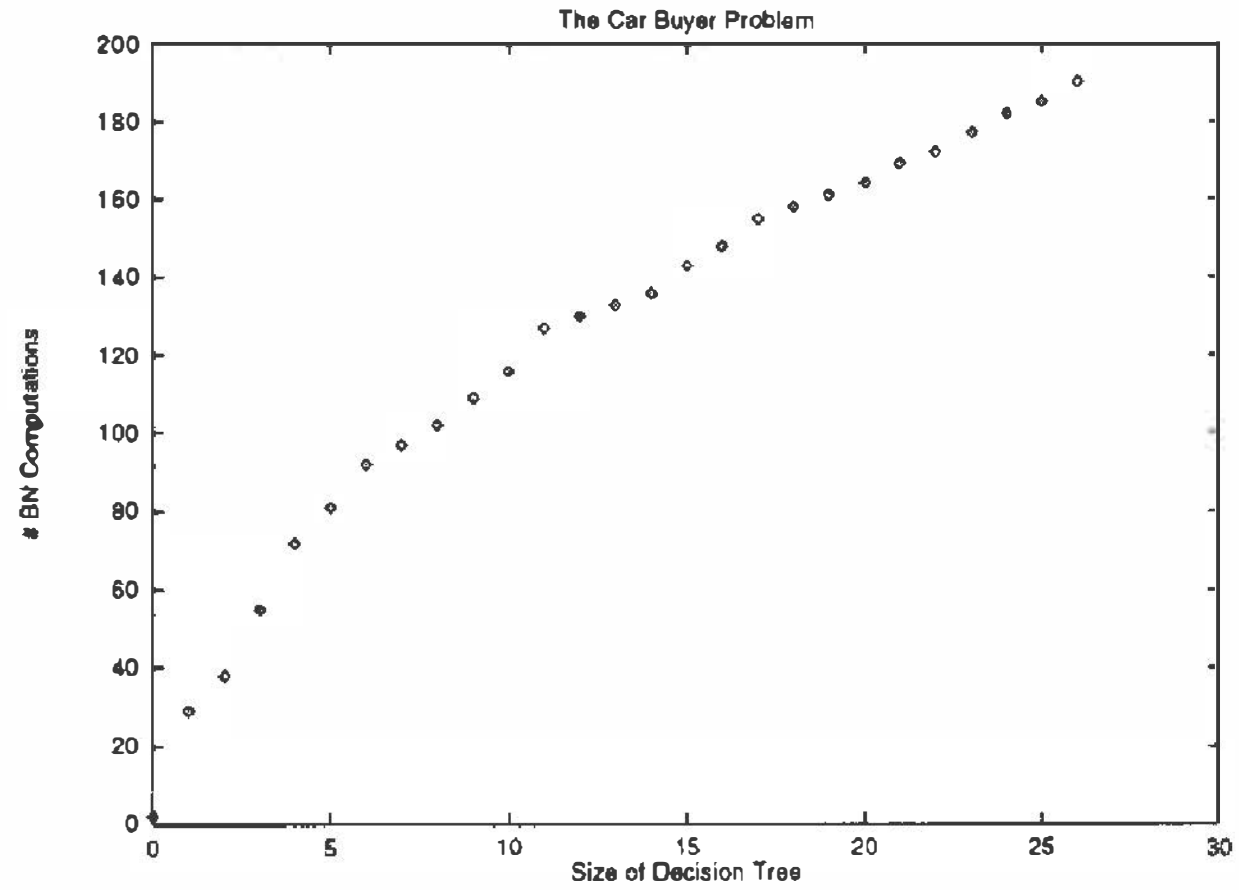

Figure 6: *Showing where the work gets done by* DT1 *when computing decision functions for the last decision in the Car Buyer Problem.*

## 5.3 PERFORMANCE ON RANDOM SINGLE DECISION IDS

To show the performance of DT1 on a single decision node, we created a number of random influence diagrams with one decision node. Figure 7 illustrates the template ID which we have used to create random decision problems. The template problem has $n$ chance nodes, each of which is an informational predecessor to the decision node. As well, each chance node is a predecessor of the value node. The template can be instantiated by choosing $n$; random binary probability distributions, *i.e.*, $P(c_k)$, are chosen from a uniform distribution on $(0, 1)$. The utility function is also chosen from a uniform distribution $[0, 1]$.

Figure 8, shows the behaviour of DT1 on seven instantiations with $n = 8$. Each point in these graphs represents a decision function; there are seven sets of data shown, one point–shape for each ID. Part (a) shows the increase in expected value as the trees increase in size. Part (b) shows the expected value of the decision function as a function of the work done by the algorithm in terms of BN calculations. The right most points represent the optimal decision tree for each problem. The left hand endpoints represent the expected value of the decision tree continuing only a single leaf.

While the data tend to overlap the general trends are clear: expected value increases with the size of the tree, and the work done. The optimal policy can be computed by the inflexible algorithm in 256 BN computations.

These data illustrate the concern mentioned in Section 3: that the greedy approach produces trees which are small but expensive to compute.

We have two points to make about these graphs. First, even the first few trees will be more valuable to a decision maker than no decision function at all, if there are deadlines or other opportunity costs. Second, the IDs in these graphs are random, and contain no asymmetries due to logical impossibilities; there is reason to believe that decision problems faced by real decision makers contain more structure than our random problems do.

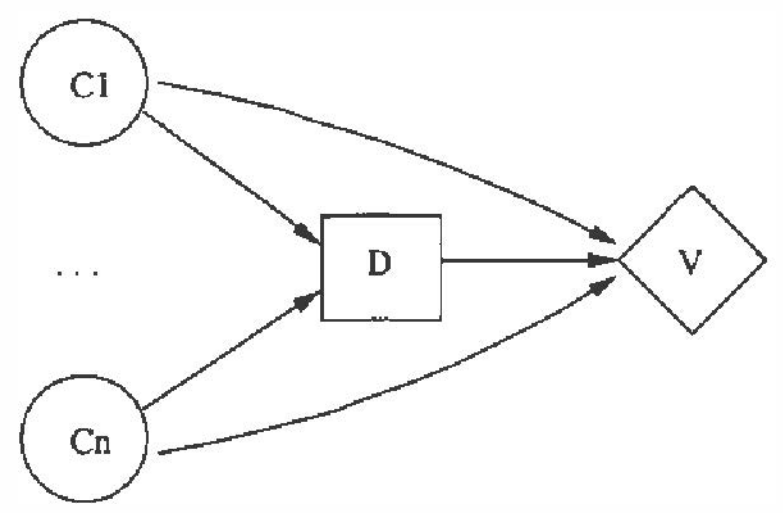

Figure 7: *An ID with one decision node and n informational predecessors.*

We implemented a random DT1, in which a leaf is chosen at random, and extended by choosing a random information variable not in its context; actions are chosen by maximizing expected utility for the randomly selected context. This program was used to find policies for the same random IDs, and Figure 9 plots the expected value versus the number of BN calculations performed. As before, the rightmost point in each set represents the complete decision tree. Because random DT1 only computes extensions it will use, there is a linear relationship between the size of the decision tree and number of BN calculations. Thus, a graph corresponding to Figure 8(a) is not presented, as its shape is identical to Figure 9.

Note that the total amount of work done by the random DT1 to complete the tree is less than half of that required by the greedy version. Also note that in the early stages of computation, the expected utility of the decision functions computed by the two versions are very similar.

## 5.4 PERFORMANCE ON IDS WITH SEVERAL DECISION NODES

We have applied the *Greedy*–DT1 technique to the Car Buyer problem. These results are preliminary, but encouraging.

Because the optimal decision tree for the last decision node can be represented with a tree of as few as 7 non-leaf vertices, we were able to explore the space of all possible decision trees for the three decision nodes. Note that the first decision node has no informational predecessors, and the second has two.

The sweep back method was able to compute the optimal expected value with a total of 33 BN calculations (compared to the 113 BN calculations required by the inflexible method). Twenty-nine of these steps were used to find a decision function for the last decision node (*Buy*), which was represented by a tree with a single internal vertex (*Result*

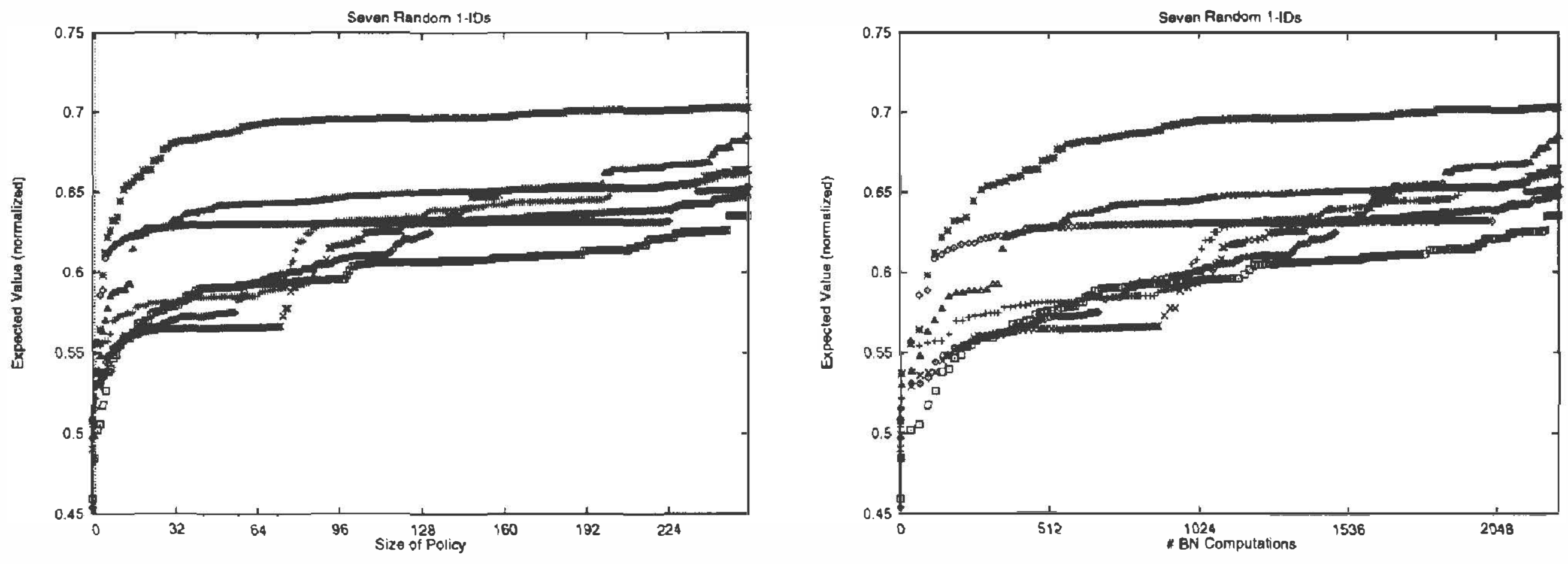

(a) (b)

Figure 8: *Seven random IDs with 8 informational predecessors each were solved by* Greedy-DT1 *with the post hoc heuristic. The improvement of the tree in terms of expected value: (a) as the tree increases in size, and (b) as the number of BN calculations performed increases.*

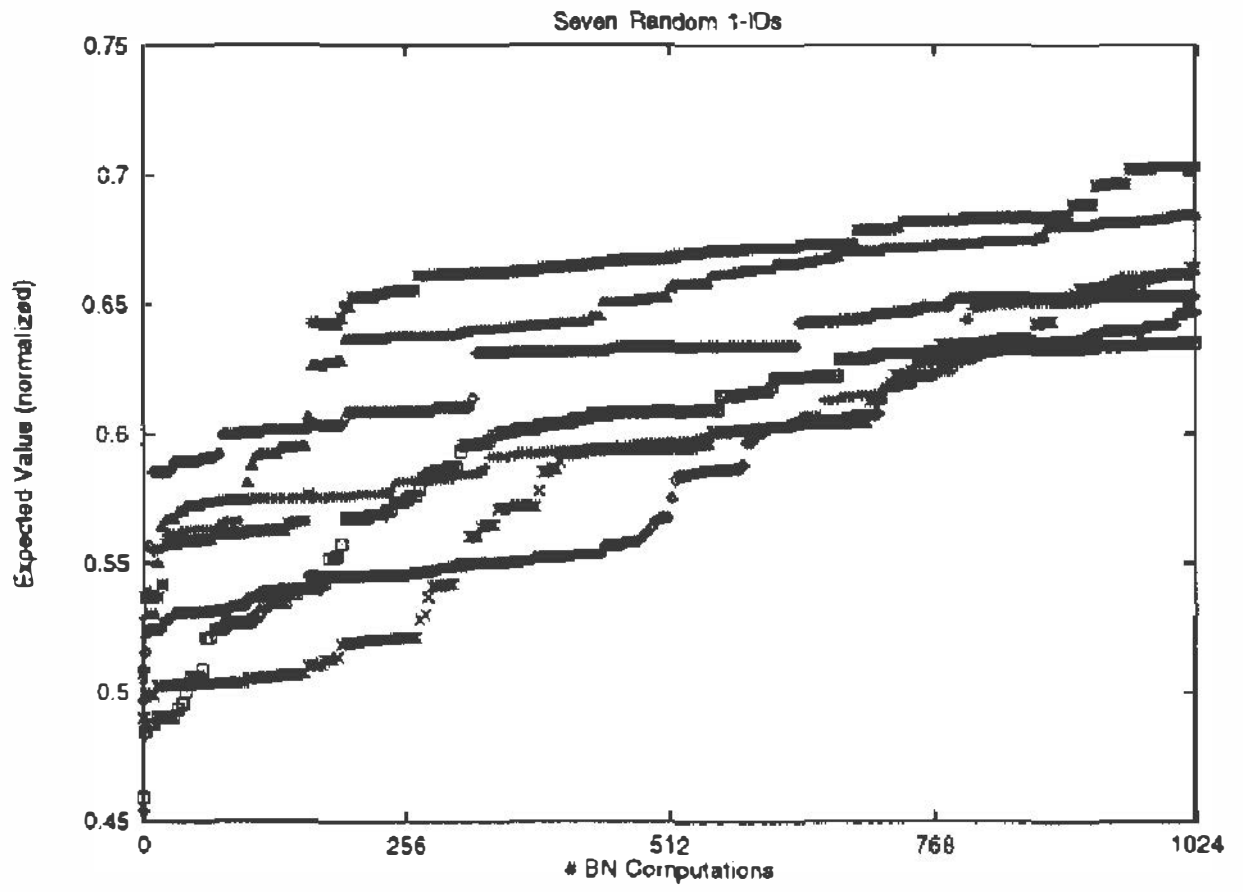

Figure 9: *The same seven random IDs as in Figure 8 each were solved by an implementation of random* DT1. *Showing the improvement of the tree in terms of expected value as the number of BN calculations performed increases.*

*of Test 1*), and four leaf vertices. The system was able to use this decision function, and its expected value to determine trivial decision trees for the remaining decision nodes, both represented by a single leaf node (an action to be performed regardless of the available information). Together, these three trees represent the optimal policy.

A sub-optimal policy, in which the decision maker decides to make no tests and buy the car is found with 6 BN calculations, and the expected value of this policy is about 3% less than the optimal policy's expected value.

# 6 CONCLUSIONS AND DISCUSSION

We have shown how a decision tree can be constructed iteratively, and that the iteration converges to the optimal decision function. Asymptotically, the number of Bayesian network calculations required by the iterative technique is a constant factor larger than dynamic programming techniques.

We have described two dimensions along which the DT1 algorithm can be varied: which leaf to extend, and how to extend a given leaf. We have presented a greedy approach to extending a leaf, and a *post hoc* heuristic for ordering the extensions. The *Greedy*–DT1 with the post hoc heuristic may not be the best use of computational resources, and we are investigating alternatives.

The greedy approach was shown to spend much effort to find the best extension to a given leaf, resulting in trees which are small but expensive to compute. The random approach, where no computational effort at all is made to determine how a context should be extended, computed much larger trees with similar computational effort. In the early stages of the process, the two versions produced decision functions whose expected value were similar. The advantage of the greedy approach is that when applied to IDs with several decision nodes, a smaller decision tree will reconnect fewer informational arcs, resulting in smaller cliques, and faster BN queries.

There are other possibilities we have not discussed in this paper, including a hyper-greedy approach, in which a leaf is extended by the first extension whose increase in expected value surpasses a given threshold.

The post hoc heuristic seems weak. However, its performance so far has not required us to seek out a better heuristic as yet. The choice of heuristic affects the value of the incremental improvement; if the heuristic does not provide a good guess as to the value of the next iteration, only a single iteration is "lost."

The DT1 algorithm is an anytime algorithm for IDs with one decision node. Each iteration has a well-defined cost, in terms of the number of BN computations required. An estimate of the increase in expected value due to a incremental improvement to the tree would provide the basis for a flexible algorithm. Our post hoc heuristic provides a means by which the previous step can be evaluated, but we are investigating this problem further.

We have shown how DT1 can be applied to IDs with several decision nodes, in the familiar sweep back technique of dynamic programming. For IDs with several decision nodes, the anytime property of DT1 is lost. We are exploring ways to balance the computational effort across the stages. The dilemma is that the decision maker may need to take action on the first decision node with some urgency, but all the computational effort could go into finding a decision function for the last decision node.

We have shown preliminary empirical results which are encouraging: small decision trees have non-trivial value to a decision maker, before inflexible techniques have produced the optimal decision tree. We take this as an encouraging result, especially towards multi-stage IDs. We have conjectured that the small trees near the end of the decision sequence provide enough information, in terms of expected value, to allow the decision maker to construct fairly detailed decision trees for decisions which must be acted upon chronologically earlier.

### Acknowledgements

The authors would like to thank the reviewers for helpful suggestions, and also Brent Boerlage of Norsys Software Corp. for advice and support in the use of the Netica API as our Bayesian network engine, and for many discussions. The second author is supported by the Institute for Robotics and Intelligent Systems, Project IC-7, and the National Sciences and Engineering Council of Canada Operating Grant OGPOO44121.